\documentclass{article}
\usepackage{ijcai24}

\usepackage{times}
\usepackage{soul}
\usepackage{url}
\usepackage[hidelinks]{hyperref}
\usepackage[utf8]{inputenc}
\usepackage[small]{caption}
\usepackage{graphicx}
\usepackage{amsmath}
\usepackage{amsthm}
\usepackage{booktabs}
\usepackage{algorithm}
\usepackage{algorithmic}
\usepackage[switch]{lineno}
\usepackage{csquotes}
\usepackage{amsmath}
\usepackage{amssymb}
\usepackage{multirow}

\title{KnowledgeNavigator: Leveraging Large Language Models for Enhanced Reasoning over Knowledge Graph}

\author{
Tiezheng Guo$^{1,2}$\and
Qingwen Yang$^{1,2}$\and
Chen Wang$^{1,2}$\and
Yanyi Liu$^{1}$ \and
Pan Li$^{1,2}$ \and
Jiawei Tang$^{1,2}$ \and
Dapeng Li$^2$ \And
Yingyou Wen $^{1,2,3,*}$
\affiliations
$^1$School of Computer Science and Engineering, Northeastern University, China\\
$^2$Neusoft AI Magic Technology Research\\
$^3$Neusoft Institute of Intelligent Medical Research
\emails
wenyingyou@mail.neu.edu.cn 
}

\begin{document}

\maketitle

\begin{abstract}
    Large language models (LLMs) have achieved outstanding performance on various downstream tasks with its advanced understanding of natural language and zero-shot capability. However, they struggle with knowledge constraints, particularly in tasks needing complex reasoning or extended logical sequences. These limitations can affect their performance in question answering (QA) by leading to inaccuracies and hallucinations. In this paper, we introduce a novel framework called KnowledgeNavigator, designed to overcome these challenges. It improves LLM reasoning by efficiently and accurately retrieving external knowledge from knowledge graph. KnowledgeNavigator starts by refining the constraints implied by the question, which helps steer the reasoning process. It then selectively gathers supporting information from the knowledge graph through a process that iteratively incorporates the LLM's insights and the question's requirements. Finally, KnowledgeNavigator prepares this structured information into LLM-friendly prompts, enhancing its reasoning capabilities. We evaluate KnowledgeNavigator on several public KGQA benchmarks, and the results demonstrate its effectiveness and generalisability. It surpasses previous methods that combine knowledge graphs with LLMs and is competitive with the fully supervised models.
\end{abstract}

\section{Introduction}

Large language models (LLMs) have gained popularity in the NLP field due to their impressive performance~\cite{touvron2023llama,anil2023palm,bai2023qwen}. LLMs pre-trained with massive data show far superior capabilities of understanding and reasoning on natural language than other language models. Although LLMs have good performance on a wide range of downstream tasks, they still struggle with knowledge-intensive challenges. Studies have shown that LLMs suffer from hallucination and knowledge limitation, including outdated or incorrect facts, and lack of specialized knowledge~\cite{zhang2023siren,martino2023knowledge,chen2023systems}. Furthermore, LLMs face difficulties in performing reasoning with long logical sequences or intricate structures~\cite{creswell2022selection}. These shortcomings restrict their use, especially in high-risk and high-sensitivity fields such as medicine.

\begin{figure}[t]
    \centering
    \includegraphics[scale=0.9]{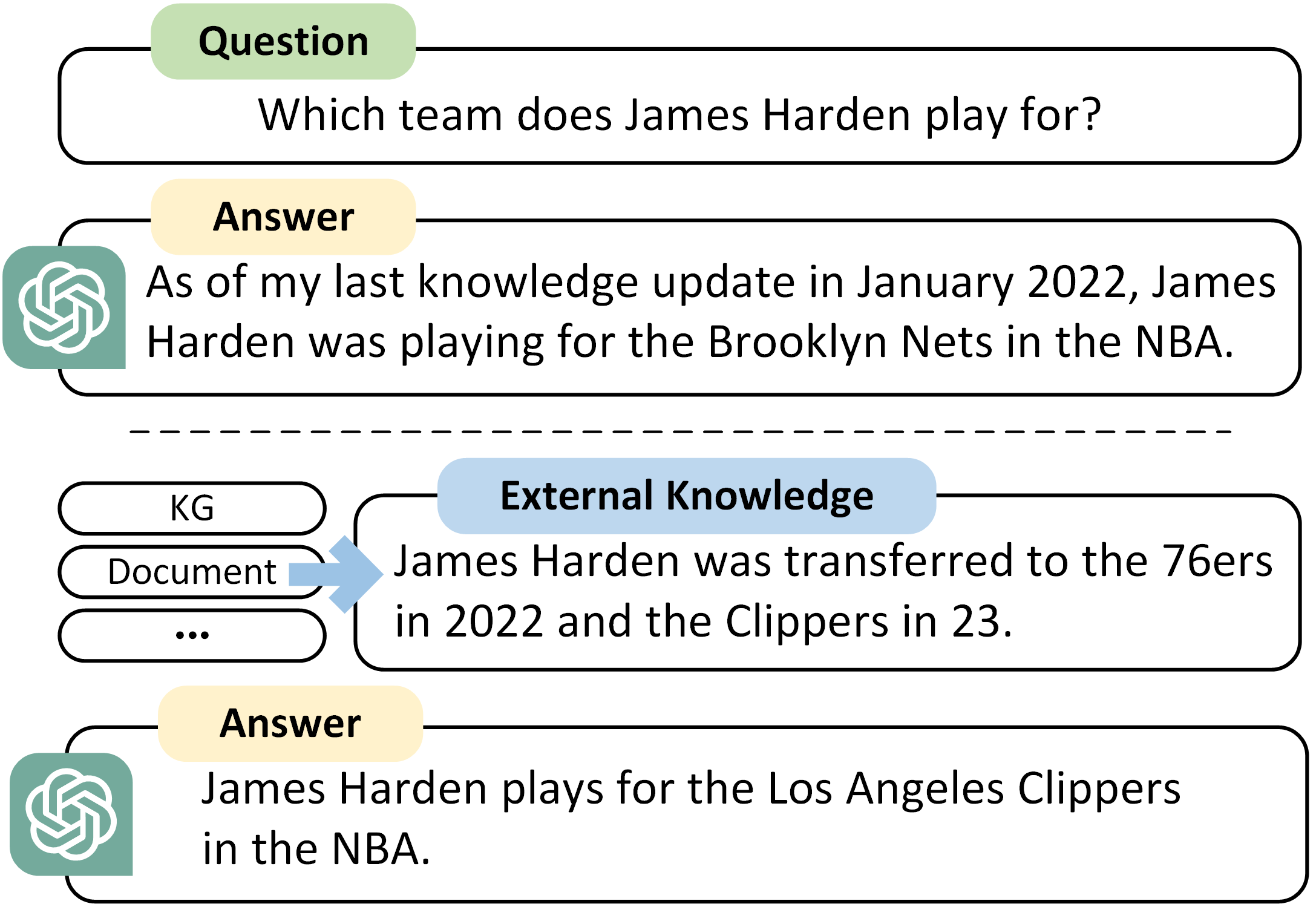}
    \caption{LLM suffers from hallucination and knowledge limitation, which can be solved with external knowledge.}
    \label{fig:intro}
\end{figure}

Improving LLMs with external knowledge is an intuitive approach to overcoming their limitations. This is particularly useful for question-answering (QA) tasks. For these tasks, the process involves retrieving correct and real-time knowledge relevant to the question, constructing prompts, and then feeding these prompts to the LLM for analysis or summary, as demonstrated in Figure~\ref{fig:intro}. A knowledge graph (KG) is a vital source of such external knowledge, which supports temporal and multimodal knowledge through structured data storage techniques~\cite{zhu2022multi,cai2022temporal}. Knowledge graphs, which store comprehensive real-world information as a graph of triples, offer more robust stronger semantic logic than plain text and are better suited to support logical reasoning tasks.

To enhance the performance of LLMs with a knowledge graph, we need to recall a multi-hop reasoning path with various entities and relations. However, there are three main challenges: First, each entity in the knowledge graph is connected to a large number of relations, but most are irrelevant to the given question. Without efficient filtering, the lengthy context and excess invalid information can lead to incorrect reasoning in LLMs~\cite{dong2023revisit}. Second, questions often require a multi-hop search over the graph, which can cause the search area to grow exponentially. Therefore, effective retrieval and pruning methods are essential. Finally, the knowledge graph's triple structure can be difficult for general LLMs to process, as they aren't typically pre-trained or fine-tuned on structured data~\cite{moiseev2022skill}. Finding an appropriate knowledge representation is crucial for effective prompting.

Considering the above challenges, we propose KnowledgeNavigator, a novel general framework to implement enhanced knowledge graph reasoning. It consists of three stages: Question Analysis, Knowledge Retrieval, and Reasoning. KnowledgeNavigator starts by predicting the retrieval scope required for the question and creates a set of similar queries. Guided by the question, it iteratively retrieves and filters relevant relations and entities at each hop within the knowledge graph. This process ensures that only necessary knowledge is recalled to answer the question. Subsequently, this knowledge is  synthesized and converted into natural language to minimize redundancy and circumvent the processing limitations of LLMs on triples. The refined knowledge is then fed to LLM for advanced reasoning. In this pipline, knowledge graph serves as an external knowledge source, while the LLM enhances the understanding of  question semantics, predicts search direction, and facilitates reasoning. Both components function as plug-ins within KnowledgeNavigator. This design allows KnowledgeNavigator to support any knowledge graph and backbone LLM, capitalizing on the timely updated knowledge and domain-specific information in the knowledge graph without the overhead of  frequent retraining of LLM.

To evaluate KnowledgeNavigator, we employ KGQA benchmarks on MetaQA and WebQSP datasets for multi-hop knowledge graph reasoning. In these datasets, KnowledgeNavigator conducts complex multi-hop reasoning on the knowledge graph with the help of LLama-2-70B-Chat and ChatGPT. KnowledgeNavigator achieves comparable performance comparing with the fully supervised models and surpasses all models that employ LLM for retrieval and reasoning. Specifically, KnowledgeNavigator outperforms KV-Mem by 16.8\%, 46.1\% and 36.8\% across the  datasets, demonstrating its effectiveness and robustness. Furthermore, we also utilize ablation studies to determine the contribution of KnowledgeNavigator's individual components to its overall performance. The findings indicate that each component is both effective and essential.

\section{Related Work}

\subsection{Knowledge Reasoning for KGQA}

Essentially, a knowledge graph is a semantic network with various entities, concepts, and relations between them~\cite{tian2022knowledge}. KGQA task as an important application of knowledge graph in the NLP field, aims to generate answers for a given question by mining and reasoning on the existing knowledge graph~\cite{chakraborty2021introduction}. Reasoning over knowledge graphs is crucial for supporting KGQA due to the inherent limitations of knowledge graphs, which can be incomplete and noisy to varying degrees. Early knowledge reasoning mainly rely on logical rules, which requires experts to design grammars and rules for specific domains. These methods have strong interpretability but require a lot of manual intervention and cannot be generalized efficiently~\cite{berant2013semantic,yih2014semantic,krotzsch2018attributed}. With the development of representation learning, many studies apply embeddings with rich semantic information to map the entities and relations to a low-dimensional vector space, and capture their potential semantic relationships to extract the optimal answer. These studies greatly improve the performance of knowledge  reasoning for KGQA, but the effectiveness of these methods relies on the representation of embedding models and lacks interpretability~\cite{huang2019knowledge,yang2014embedding,liu2017endo}. To better solve the complex multi-hop reasoning, more researchers currently apply neural networks to learn the interaction patterns between entities and relations in the knowledge graph to achieve automatic and accurate reasoning, and improve the generalization of reasoning models~\cite{xiong2019improving,das2017go}.

\subsection{Knowledge Graph Enhanced LLM}

Knowledge graph supports the structured representation of real-world knowledge, through temporal and personalized design, and can meet a variety of different knowledge storage and usage requirements~\cite{ji2021survey}. Therefore, knowledge graph is applied to enhance LLM pre-training and LLM generation as an important knowledge source~\cite{pan2023unifying}. The knowledge graph contains structured information that has clearer logic and reasoning paths compared to natural language. Therefore, many studies utilize entities and relations to build a corpus and design various training tasks, aiming to enhance the effectiveness of LLM pre-training~\cite{zhang2019ernie,feng2023factkb,yu2022jaket}. However, both retraining and continue pretraining of LLM require high computing resources and time costs, making it challenging to keep up with the rapidly evolving knowledge applications. Therefore, a more straightforward approach to address the lack of  knowledge in LLM is to construct knowledge-enhanced prompts with factual information. Many works retrieve knowledge related to the target question through external retrieval algorithms, and incorporate this knowledge into prompts for LLM, enabling it to reason in unfamiliar domains~\cite{baek2023knowledge,wu2023retrieve,ji2022rho}. In this paper, we also adopt the method of enhancing LLM generation with knowledge graph to improve its performance in KGQA tasks.

\section{Method}

\begin{figure*}[t]
    \centering
    \includegraphics[scale=0.85]{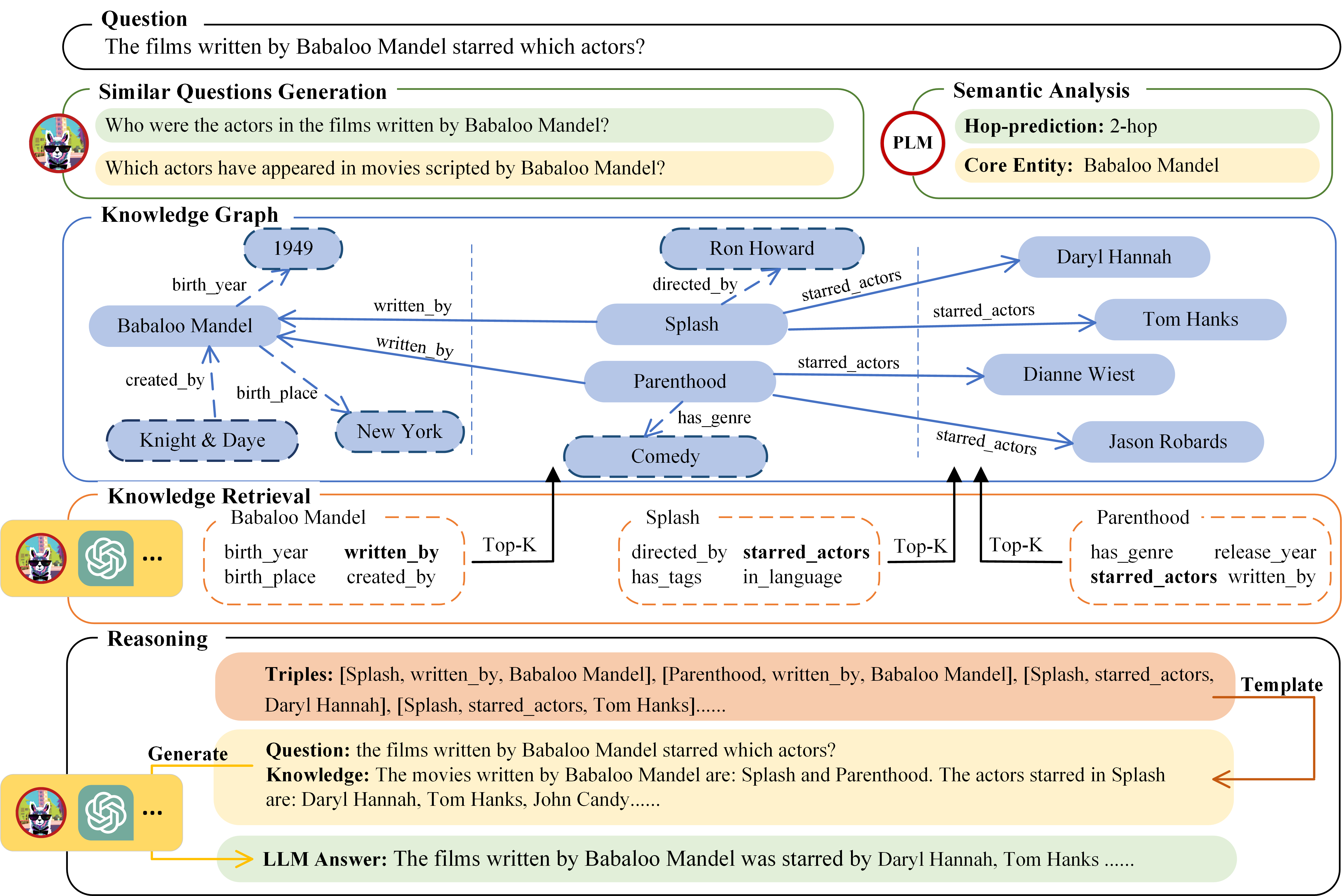}
    \caption{An overview of KnowledgeNavigator. The framework consists of three consecutive phases: Question Analysis, Knowledge Retrieval, and Reasoning. The given example comes from MetaQA, describing a 2-hop reasoning task starting from \enquote{Babaloo Mandel} and ending with entities including \enquote{Tom Hanks}. In the knowledge graph, solid lines indicate that entities or relations are retrieved as reasoning knowledge, while dashed lines indicate that entities or relations are discarded.}
    \label{fig:structure}
\end{figure*}

We design KnowledgeNavigator to support KGQA tasks by performing enhanced reasoning on knowledge graph. The reasoning process of KnowledgeNavigator contains three stages: Question Analysis, Knowledge Retrieval, and Reasoning as shown in Figure~\ref{fig:structure}.

\subsection{Question Analysis}

The multi-hop reasoning of question is the main challenge in KGQA tasks. The Question Analysis stage enhances and restricts the reasoning through pre-analyzing the given question, which supports enhanced reasoning on the knowledge graph. This approach helps to improve the retrieval efficiency and accuracy.

To answer a question $Q$, KnowledgeNavigator first predicts the potential hop number $h_Q$ of the question to obtain all the knowledge required to it starting from the core entities. The hop number indicates the maximum reasoning depth required to retrieve the information. The process of hop number prediction is a classification task. KnowledgeNavigator implements it with a fine-tuned pre-trained language model (PLM) and a simple linear classifier:

\begin{equation}
    V_Q = PLM(Q)
\end{equation}

\begin{equation}
    h_Q = \arg\max_{h} P(h|V_Q), h \in \{1, 2, \ldots, H\}
\end{equation}

The reasoning logic of each question in KGQA task is implied in the semantics of the question itself. Therefore knowledge graph reasoning is a process of mining this reasoning logic from the question. In order to enhance this mining, KnowledgeNavigator generates a set of similar questions $S = {\{s^Q_1, s^Q_2, \ldots, s^Q_m\}}$ with the same semantics as the original question, using LLM. Various ways of phrasing the same question can shed light on the reasoning logic from different angles. Therefore, these similar questions serve to enrich the information available during the Knowledge Retrieval stage.

In the case of Figure~\ref{fig:structure}, KnowledgeNavigator predicts the number of reasoning hops $h_Q$ starting from the core entity \enquote{Babaloo Mandel} is 2 with PLM fine-tuned with the MetaQA 2-hop dataset. It then generates $S$ containing two variants of the original question.

\subsection{Knowledge Retrieval}

Extracting relevant knowledge from the knowledge graph is crucial for answering a given question. The Knowledge Retrieval stage aims to extract the logical path by performing advanced reasoning on the knowledge graph. This constructs a smaller, focused subgraph that aids in generating answers. The retrieval process is mainly achieved by interacting with LLM, which helps avoid the expense of retraining the model for various tasks.

Knowledge Retrieval is an iterative search process with a depth limit of $h_Q$. In each iteration $i$, KnowledgeNavigator begins with a set of core entities $E_i = {\{e^1_i, e^2_i, \ldots, e^n_i\}}$. It then explores all one-hop relations connected to each entity, forming a candidate relation set $R^n_i = \{r^{n,1}_i, r^{n,2}_i, \ldots, r^{n,k}_i\}$. Since an entity may have many relations in a knowledge graph, not all are relevant to the question. It is necessary to prune the reasoning path to minimize the influence of unrelated or noisy knowledge on answer generation. KnowledgeNavigator linears the candidate relations for each entity into a string and format it alongside the entity and question variations in $S$ as prompts for LLM. The LLM is tasked with choosing the $K$ most relevant relations from $R^n_i$ based on the question variant.

Based on the results of relation filtering, a weighted voting mechanism is employed to rank the frequency of each relation linked to entity $e^n_i$. Relations chosen for the original question are given double the weight of those from variants generated by the LLM in the first stage:

\begin{equation}
    \text{Score}(r) = \sum_{s \in S} w(s) \cdot \mathbb{I}(r, LLM(e, s, R))
\end{equation}

\begin{equation}
    w(s) = \begin{cases}
        2 & \text{if } s = Q \\
        1 & \text{otherwise}
        \end{cases}
\end{equation}

At this stage, the indicator function $\mathbb{I}$ denotes whether a relation $r$ from set $R$ is chosen by the LLM. Specifically, the function assigns a value of 1 if the relation is selected and 0 if it is not.

The ranking of relations for each entity is carried out independently, enhancing the diversity of the reasoning process. After filtering all relations in iteration $i$, KnowledgeNavigator selects $M$ optimal relations for each entity. It then queries the triples $(e^n_i, optimal\_r^n_i, tail)$ and $(head, optimal\_r^n_i, e^n_i)$ from knowledge graph. These triples form part of the reasoning path and are included in the retrieved knowledge set $RK$. The untraversed entities in $tail$ and $head$ are compiled into the core entity set $E_{i+1}$ for the next iteration. 

In this stage, KnowledgeNavigator begins with the core entities $E_0$ extracted from the given question $Q$. It then iteratively filters relations and incorporates triples to $RK$ untils $h_Q$ is reached. These triples in $RK$ will be used as reasoning knowledge for the next stage.

\subsection{Reasoning}

Through several iterations, KnowledgeNavigator summarizes enough knowledge in $RK$ to address the given question. The Reasoning stage then leverages this knowledge to generate the answer.

The knowledge retrieved from knowledge graph is structured as triples in the format of $[head, relation, tail]$. Each triple is an implicit expression of the reasoning path. To fully answer a question, we can link the entities and relations from multiple triples to create a reasoning path and further a reasoning sub-graph. By merging nodes and condensing this sub-graph through triple aggregation, we can enhance the reasoning efficiency of the LLMs. For instance, KnowledgeNavigator aggregates triples $T$ within $RK$ that share the same head or tail entity and relation into a single, consolidated triple:

\begin{equation}
    f_{head}(T) = \left\{ (h, r, [a_1, \ldots, a_n]) \mid \forall (h, r, a_i) \in T \right\}
\end{equation}

\begin{equation}
    f_{tail}(T) = \left\{ ([h_1, \ldots, h_n], r, a) \mid \forall (h_i, r, a) \in T \right\}
\end{equation}

This can effectively reduce redundant information details and enhances the ability to represent knowledge. KnowledgeNavigator then convert the aggregated triples into natural language using templates (\textit{e.g.~The~\{relation\}~of~\{head\}~is(are):~\{tail\}}). This can circumvent the limited capacity of LLM to understand data structured as triples. Subsequently, the natural language-formatted knowledge is merged into a single string and fed to LLM along with the question $Q$. The LLM is prompted to generate an answer completely based on the provided external knowledge without using its own learned knowledge.

\section{Experiments}
\subsection{Dataset}

To test the ability of KnowledgeNavigator on multi-hop knowledge graph reasoning tasks, we evaluate it on two datasets: MetaQA~\cite{MetaQA} and WebQSP~\cite{WebQSP}. In the KGQA task on both datasets, we use Hits@1 as the evaluation metric to evaluate the correctness of the answers generated by LLM.

MetaQA is a large-scale KGQA dataset in the movie domain, which provides a knowledge graph with 43k entities, 9 relations and 135k triples, as well as 407k questions. The question set is extracted from the Facebook MovieQA dataset, containing questions that require 1-hop to 3-hop reasoning away from the head entities. Each question  consists of a head entity, a relation reasoning path, and the answer entities. To verify KnowledgeNavigator's multi-hop reasoning capability, we specifically use the 2-hop and 3-hop vanilla datasets in MetaQA for experiment.

WebQSP is a benchmark with fewer questions but a large-scale knowledge graph, which can effectively evaluates the large-scale search ability of KnowledgeNavigator. WebQSP provides questions up to 2 hops based on freebase, each question contains a topic entity, constraints, inferential chains and SPARQL queries for finding the answer. We set up the knowledge graph with the latest version of freebase data dumps provided by google including 3.12B triples~\cite{freebasedatadumps}. WebQSP provides 4737 questions and we remove 11 questions which have no gold answers from it. 

\subsection{Baselines}

To evaluate the effectiveness of KnowledgeNavigator in reasoning on knowledge graphs, we compare it with a set of well-known baseline models in the field of KGQA, which are all built with fully supervised. These baselines include KV-Mem~\cite{KVMem}, GraftNet~\cite{GraftNet}, EmbedKGQA~\cite{EmbedKGQA}, NSM~\cite{NSM}, UniKGQA~\cite{UniKGQA}. All of these baselines were evaluated on both MetaQA and WebQSP. In addition, we add StructGPT~\cite{structgpt} and TOG~\cite{TOG} as baselines of using LLM for KGQA tasks. These two frameworks are both based on un-fine-tuned LLM for knowledge retrieval and question reasoning. We use the results of these two frameworks using the same LLM as KnowledgeNavigator as a reference. It is important to note that TOG was only tested on WebQSP. 

We apply the LLama-2-70B-Chat~\cite{llama2} and ChatGPT as large language model baselines. Specifically, we deploy the same template to prompt the large language models. The only distinction between this baseline and KnowledgeNavigator is the external knowledge retrieved.

\subsection{Experiment Details}

KnowledgeNavigator is decoupled from LLM, any LLM can be used as a plug-in component for reasoning. We use ChatGPT and LLama-2-70B-Chat as the LLM component in experiments. We call ChatGPT with the OpenAI API. LLama-2-70B-Chat is deployed locally with 4 NVIDIA A100 80G with vllm framework~\cite{vllm}. The context length is set to 4096 as default, and the maximum number of tokens to generate per output sequence is set to 1024.  We fine-tune bert-base-uncased and a linear classifier on the training set of the datasets for hop prediction.

For both datasets, KnowledgeNavigator generates two variants for each question, and the hop prediction is conducted within the range of 1 to 3. In MetaQA, KnowledgeNavigator performs weighted ranking on the top-1 relation for each (question~variant, entity, relations) group and selects the top-1 ranked result for the next iteration. For WebQSP, these two parameters are set to top-2. In the few-shot scenario, few-shot is composed of two examples from the training set of the same dataset, which is in the same format as the target task.

\subsection{Main Results}
\begin{table}
    \centering
    \begin{tabular}{lrrr}
        \toprule
        \multirow{2}{*}{Methods} & \multicolumn{1}{c}{MQA} & \multicolumn{1}{c}{MQA} & \multirow{2}{*}{WebQSP} \\
        & \multicolumn{1}{c}{2-hop} & \multicolumn{1}{c}{3-hop} & \\
        \midrule
        KV-Mem     & 82.7    & 48.9 & 46.7 \\
        GraftNet   & 94.8    & 77.7 & 66.4 \\
        EmbedKGQA  & 98.8    & 94.8 & 66.6 \\
        NSM        & \textbf{99.9}    & 98.9 & 68.7 \\
        UniKGQA    & 99.0    & \textbf{99.1} & 75.1 \\
        TOG~(LLama-2-70B-Chat)  & -       & -    & 68.9 \\
        TOG~(ChatGPT)  & -       & -    & \textbf{76.2} \\
        StructGPT(ChatGPT)    & 97.3    & 87.0 & 72.6 \\
        \midrule
        LLama-2-70B-Chat   & 34.7  & 35.6  & 55.8 \\ 
        ~+ ours            & 98.9  & 77.5  & 71.8 \\
        ~+ ours~(fewshot)  & \textbf{99.5} & 87.8  & 77.0 \\
        ChatGPT            & 31.0  & 43.2   & 61.2  \\
        ~+ ours            & 96.3  & 77.2  & \textbf{83.5}  \\
        ~+ ours~(fewshot)  & 99.1  & \textbf{95.0}  & 82.3  \\
        \bottomrule
    \end{tabular}
    \caption{The performance of KnowledgeNavigator and baselines on MetaQA and WebQSP. The best result in each block is in bold.}
    \label{tab:main result}
\end{table}

Table~\ref{tab:main result} shows the performance of KnowledgeNavigator and the baselines on KGQA datasets. KnowledgeNavigator achieves impressive accuracy of 99.5\% on MetaQA 2-hop with LLama-2-70B-Chat, 95.0\% and 83.5\% on MetaQA 3-hop and WebQSP task with ChatGPT.

First, LLM can answer questions in KGQA tasks without relaying on the external knowledge, and even outperforms KV-Mem on the WebQSP benchmark. This demonstrates that the LLM's pre-training knowledge is useful for answering questions that have not been learned directly. Nonetheless, there is still a significant performance gap between the LLM and state-of-the-art models in KGQA tasks. This suggests that the LLM faces challenges in reasoning and answering complex questions using only its internal knowledge.

We found that KnowledgeNavigator can enhance the accuracy of LLMs in reasoning and question answering tasks. By employing KnowledgeNavigator's search and prompting capabilities, LLMs achieve significant improvements on KGQA. In addition, the performance of KnowledgeNavigator can be further improved by adding a few incontext examples. In a comparison with all baseline models, KnowledgeNavigator achieves best results on WebQSP. KnowledgeNavigator surpasses other research using the same backbone LLM for knowledge retrieval and reasoning, such as TOG and StructGPT, on MetaQA. Even against fully supervised models, KnowledgeNavigator can also achieve competitive performance on MetaQA. KnowledgeNavigator exhibits improvements over KV-Mem with increases of 16.8\%, 46.1\% and 36.8\% respectively on the three datasets. However, KnowledgeNavigator's performance on MetaQA 3-hop questions is inferior to NSM and UniKGQA. This is due to the complex semantics of multi-hop questions which bring challenges to relation filtering, and suggest that the deep reasoning capability of LLM still needs to be further enhanced.

\subsection{Ablation Study}

\begin{table}
    \centering
    \begin{tabular}{lrrr}
        \toprule
        \multirow{2}{*}{Similar questions} & \multicolumn{1}{c}{MQA} & \multicolumn{1}{c}{MQA} & \multirow{2}{*}{WebQSP} \\
        & \multicolumn{1}{c}{2-hop} & \multicolumn{1}{c}{3-hop} & \\
        \midrule 
        KnowledgeNavigator \\
        ~- w/o similar question    & 99.4    & 86.4 & 72.6 \\
        ~- w/ 2 similar questions  & 99.5    & 87.8 & \textbf{77.0} \\
        ~- w/ 4 similar questions  & \textbf{99.9}  &  \textbf{88.3}    &  74.2    \\
        \bottomrule
    \end{tabular}
    \caption{Performance of KnowledgeNavigator with different number of similar questions on MetaQA and WebQSP.}
    \label{tab:Ablation study similar}
\end{table}

\begin{table}
    \centering
    \begin{tabular}{lrrr}
        \toprule
        \multirow{2}{*}{Knowledge Format} & \multicolumn{1}{c}{MQA} & \multicolumn{1}{c}{MQA} & \multirow{2}{*}{WebQSP} \\
        & \multicolumn{1}{c}{2-hop} & \multicolumn{1}{c}{3-hop} & \\
        \midrule
        KnowledgeNavigator \\
        ~- w/ single triples        & 99.1    & 83.6 & 70.6 \\
        ~- w/ joined triples        & 99.4    & 87.4  & 73.9 \\
        ~- w/ single sentences   & 99.5    & 87.0 & 75.0 \\
        ~- w/ joined sentences   & \textbf{99.5} & \textbf{87.8} & \textbf{77.0} \\
        \bottomrule
    \end{tabular}
    \caption{Performance of KnowledgeNavigator with different knowledge formats on MetaQA and WebQSP.}
    \label{tab:Ablation study knowledge}
\end{table}

We perform an ablation study on KnowledgeNavigator, aiming to analyze the impact of similar questions and knowledge representation forms. The ablation study involves experiments with varying number of similar questions and forms of knowledge representation. In the ablation study, we use LLama-2-70B-Chat as the backbone LLM and use the same prompt template with 2-shot examples for all cases. Table~\ref{tab:Ablation study similar} and \ref{tab:Ablation study knowledge} shows the results of the ablation study.

\subsubsection{Impact of Number of Similar Questions}

Table~\ref{tab:Ablation study similar} shows the accuracy of using 0, 2, and 4 similar questions for voting in relation selection. For MetaQA, the performance of KnowledgeNavigator increases slowly with the number of similar questions, since more voters make the relation selection more stable. For WebQSP, the low quality of the original questions makes it difficult for LLM to generate similar questions with identical meanings.  Consequently, these similar questions have limited help to KnowledgeNavigator and may lead to voting errors. Moreover, as the number of KnowledgeNavigator's requests to the LLM rise linearly with the number of similar questions, there's a balance to be struck between computational costs and it's effectiveness. In our experiments, we use 2 similar questions as default to control the computational cost.

\subsubsection{Impact of Knowledge Formats}

Table~\ref{tab:Ablation study knowledge} shows the impact of different knowledge representation forms on the performance of KnowledgeNavigator. In this part, we use the same knowledge with different representation forms to prompt LLM. Specifically, for \textquote{w/ single triples} and \textquote{w/ joined triples}, all triples are concatenated into a string in the form of $[head, relation, tail]$ or $[head, relation, [tail_1, \ldots, tail_n]]$, for \textquote{w/ single sentences}, each triple is converted into a separate natural language sentence using a template and concatenated into a string.

For different knowledge formats, we found that the performance of KnowledgeNavigator increases with the logical closeness of the knowledge representation. First, for both triples and sentences, using aggregated knowledge can effectively reduce the redundant information and improve the density of the knowledge in prompt, therefore reduce the difficulty of reasoning. Second, for the general LLMs that haven't been fine-tuned, using knowledge in natural language format can avoid the errors caused by their insufficient understanding capabilities on structured data.

\subsection{Error Analysis}

\begin{table}
    \centering
    \begin{tabular}{lrrr}
        \toprule
        \multirow{2}{*}{Error type} & \multicolumn{1}{c}{MQA} & \multicolumn{1}{c}{MQA} & \multirow{2}{*}{WebQSP} \\
        & \multicolumn{1}{c}{2-hop} & \multicolumn{1}{c}{3-hop} & \\
        \midrule
        Relation Selection Error & 6    & 95 & 69 \\
        Reasoning Error         & 79   & 5  & 10 \\
        Hallucination           & 5    & 0  & 4 \\
        other Errors           & 10   & 0  & 17 \\
        \bottomrule
    \end{tabular}
    \caption{Distribution of random 100 error samples on each dataset.}
    \label{tab:Error Analysis}
\end{table}

To analyze the causes of errors in KnowledgeNavigator, we randomly extract 100 error samples from the results of MetaQA and WebQSP. The errors are manually analyzed and classified into 4 categories according to the reasons:

\begin{enumerate}
    \item Relation Selection Error: Wrong relations are selected in the Knowledge Retrieval stage, resulting in the failure to retrieve correct knowledge.
    \item Reasoning Error: KnowledgeNavigator retrieves the correct knowledge, but performs wrong reasoning in answer generating.
    \item Hallucination: KnowledgeNavigator does not generate answers based on the retrieved external knowledge.
    \item Other Errors: Including intermediate errors causing search interruption, and excessively long context leading to knowledge truncation, etc.
\end{enumerate}

Table~\ref{tab:Error Analysis} shows the error analysis results on the three datasets. It is easy to find that the error distribution on the three datasets are different. Furthermore, the reasoning error is the main error on MetaQA 2-hop and the relation selection error appears most on MetaQA 3-hop and WebQSP. This is because the semantics of the questions in MetaQA 3-hop and WebQSP are more complex: Question in MetaQA 3-hop features a longer reasoning path and a more intricate knowledge sub-graph, challenging the relation reasoning capabilities of LLM. 
For WebQSP, each entity is associated with numerous similar relations in freebase, which complicates the task for LLM to select the high relevant relations for the next iteration. Meanwhile, as the reasoning logic of questions in MetaQA 2-hop is more straightforward, LLM rarely selects the wrong relations, but shows insufficient reasoning ability.

According to the error statistics, we can improve the performance of LLM on KGQA tasks through targeted optimization. Specifically, we can improve its ability of selecting the relevant relations by enhancing the semantics of the questions or strengthening the connecting between the knowledge graph and the reasoning path. We can also help LLM to perform reasoning by optimizing the prompt and knowledge representation. We will keep working on these issues in the future.

\subsection{Case Study}

Figure~\ref{fig:structure} is an example of KnowledgeNavigator performing a KGQA task. First of all, KnowledgeNavigator predicts the reasoning hop number starting from the core entity \enquote{Babaloo Mandel} based on a PLM, and generates 2 similar questions for the target question with LLM. 

Then, KnowledgeNavigator extracts all relations linked to \enquote{Babaloo Mandel} and serializes them into \enquote{birth\_year;~birth\_place;~wrritten\_by;~created\_by}, as part of the prompt. For the only core entity \enquote{Babaloo Mandel}, LLM selects the optimal relation linked to it based on each question variant and gets the weighted voting result \{written\_by:3, created\_by:1\}. As the number of optimal relation selection for each entity is set to one, the triples with \enquote{written\_by} as the optimal relation and \enquote{Babaloo Mandel} as head or tail entity (\textit{i.e.~[Splash, written\_by, Babaloo Mandel] and [Parenthood, written\_by, Babaloo Mandel]}) are extracted as the first step of the reasoning path. The tail entities \enquote{Splash} and \enquote{Parenthood} are selected as the core entities for the second iteration to continue the knowledge retrieval.

After KnowledgeNavigator reached the predicted hops, the triples retrieved from the knowledge graph can be combined into an effective reasoning path (\textit{e.g.~Babaloo Mandel - written\_by - Splash - starred\_actors - Dianne Wiest}), and further into a reasoning sub-graph. Take the knowledge about \enquote{Splash} as example, KnowledgeNavigator will first combine the triples into \enquote{[Splash, startted\_actors, [Dary Hannah, Tom Hanks]]}, and then convert it into \enquote{The actors starred in Splash are: Dary Hannah and Tom Hanks} with template. All triples retrieved are finally concatenated into a string and feed to LLM as part of the answer generation prompt.

\section{Conclusion}

In this paper, we study the challenge of knowledge limitations in LLM and introduce a framework KnowledgeNavigator to improve the reasoning and question answering capabilities of LLM on knowledge graph. KnowledgeNavigator consists of three stages: Question Analysis, Knowledge Retrieval and Reasoning. During Question Analysis, KnowledgeNavigator first pre-analyzes the question and generates variants for it to assist reasoning. Then, relying on the guidance of LLM, it iteratively retrieves and filters candidate entities and relations within the knowledge graph to extracts relevant external knowledge. Finally, this knowledge is transformed into an effective prompt to improve LLM's performance on knowledge-intensive tasks. We evaluate KnowledgeNavigator with KGQA metrics, and the results indicate that introducing external knowledge from the knowledge graph benefits LLM in handling complex tasks. KnowledgeNavigator outperforms other frameworks deploying LLM on enhanced KGQA and achieves comparable performance with previous fully supervised models. We also conduct an ablation study to confirm the effectiveness of each component in KnowledgeNavigator and analyze the errors to identify the further research directions.

\section{Limitations}

Although KnowledgeNavigator has shown impressive performance on KGQA tasks, it still has some limitations. First, the logic path extraction and knowledge reasoning are achieved by feeding prompts built with multiple sentences to LLM. Therefore, the performance of KnowledgeNavigator is closely related to the natural language understanding and reasoning ability of the backbone LLM. It might be challenging to apply KnowledgeNavigator to LLMs with fewer parameters or shorter context lengths. Second, the answer generation process of KnowledgeNavigator contains multiple interactions with LLM. This can result in longer delays, especially in questions with complex reasoning paths, and may not be suitable for scenarios that have strict response time requirements.  Therefore, further research should focus on these limitations to improve the application of LLM in KGQA.

\section*{Acknowledgments}
This study was supported by Liaoning Provincial Science and Technology Innovation Project in the Field of Artificial Intelligence (Project name:Research on key technologies for systems engineering of large language model) and in part by Grants from Shenyang Science and Technology Plan Project (Grant No. RC210469). 

\bibliographystyle{named}
\bibliography{ijcai24}

\end{document}